\documentclass{article}

\usepackage[preprint]{neurips_2026}

\usepackage[utf8]{inputenc}
\usepackage[T1]{fontenc}
\usepackage{hyperref}
\usepackage{url}
\usepackage{booktabs}
\usepackage{amsfonts}
\usepackage{amsmath}
\usepackage{amssymb}
\usepackage{nicefrac}
\usepackage{microtype}
\usepackage{xcolor}
\usepackage{graphicx}
\usepackage{subcaption}
\usepackage{algorithm}
\usepackage{algpseudocode}
\usepackage{multirow}
\usepackage{pifont}
\usepackage{tcolorbox}

\newcommand{\halfmark}{\rlap{\checkmark}\hspace{0.05em}\rule[0.55ex]{0.85em}{0.4pt}}

\title{ResearchEVO: An End-to-End Framework for\\Automated Scientific Discovery and Documentation}

\author{%
  Zhe Zhao, Haibin Wen, Jiaming Ma, Jiachang Zhan, Tianyi Xu, Ye Wei, Qingfu Zhang\\
  City University of Hong Kong, Hong Kong, 999077, China
}

\begin{document}

\maketitle

\begin{abstract}
An important recurring pattern in scientific breakthroughs is a two-stage process: an initial phase of undirected experimentation that yields an unexpected finding, followed by a retrospective phase that explains why the finding works and situates it within existing theory.
We present \textbf{ResearchEVO}, an end-to-end framework that computationally instantiates this discover-then-explain paradigm.
The \textbf{Evolution Phase} employs LLM-guided bi-dimensional co-evolution---simultaneously optimizing both algorithmic logic and overall architecture---to search the space of code implementations purely by fitness, without requiring any understanding of the solutions it produces.
The \textbf{Writing Phase} then takes the best-performing algorithm and autonomously generates a complete, publication-ready research paper through sentence-level retrieval-augmented generation with explicit anti-hallucination verification and automated experiment design.
To our knowledge, ResearchEVO is the first system to cover this full pipeline end to end: no prior work jointly performs principled algorithm evolution and literature-grounded scientific documentation.
We validate the framework on two cross-disciplinary scientific problems---Quantum Error Correction using real Google quantum hardware data, and Physics-Informed Neural Networks---where the Evolution Phase discovered human-interpretable algorithmic mechanisms that had not been previously proposed in the respective domain literatures.
In both cases, the Writing Phase autonomously produced compilable \LaTeX\ manuscripts that correctly grounded these blind discoveries in existing theory via RAG, with zero fabricated citations.
\end{abstract}

\section{Introduction}
\label{sec:intro}

Many of the most impactful scientific discoveries have unfolded in two stages. Mendel crossed thirty thousand pea plants before formulating the laws of heredity. Curie systematically measured hundreds of minerals before proposing the theory of radioactivity. Newton computed for two decades before writing the Principia. Darwin collected specimens across the Galapagos for five years, then spent twenty-three more years distilling his observations into a coherent theory of natural selection. In every case, the unglamorous work of experimentation came first; the celebrated theoretical insight followed as a retrospective explanation. This recurring pattern reveals an important structure underlying productive scientific inquiry, which we term the discover-then-explain cycle. The first stage---discovery---is characterized by divergent, unconstrained exploration driven solely by empirical feedback: does this work or not? The second stage---explanation---is characterized by convergent, literature-grounded analysis that connects raw findings to existing theory and organizes them into communicable knowledge. The two stages serve complementary cognitive functions that are difficult to execute simultaneously: a scientist who must explain while still experimenting will tend to pursue only those experiments that are easy to explain, potentially missing the most important but least intuitive discoveries.

With the advancement of artificial intelligence, both stages of this cycle have been individually automated to varying degrees---but never jointly. On the discovery side, early work on automated machine learning demonstrated that entire learning algorithms could be evolved from mathematical primitives~\cite{automl_zero}, and neural architecture search~\cite{nas_survey} showed that network topologies could be discovered without human intervention. More recently, large language models~\cite{openai2023gpt4} have opened a qualitatively new frontier: rather than searching over numerical parameters or graph structures, LLMs can directly generate, mutate, and recombine executable code, effectively serving as intelligent evolutionary operators over the space of algorithms~\cite{funsearch,liu2024evolution,reevo,alphaevolve,llamea}. These LLM-guided evolution systems have achieved remarkable results, from discovering novel mathematical solutions to producing the first improvement over a classical matrix multiplication algorithm in over fifty years. Yet they share two fundamental limitations. On the process side, most methods confine their search to a single dimension---evolving the logic of a fixed function within a predefined template---while leaving the overall algorithm architecture untouched. This restriction caps the achievable innovation at incremental refinements rather than structural breakthroughs. On the output side, every existing evolution system terminates at a code artifact: an optimized function or program with no accompanying scientific narrative, no connection to existing theory, and no explanation of why the discovered solution works. In other words, these systems automate the first stage of the scientific cycle---discovery---but leave the second stage---explanation---entirely to humans.

On the explanation side, a parallel line of research has tackled the problem of automated scientific writing. Systems for end-to-end research automation~\cite{ai_scientist,ai_scientist_v2} can generate ideas, execute experiments, and produce complete research papers. Research-review closed loops~\cite{cycleresearcher} iteratively refine paper quality through preference-optimized reviewer models. Multi-agent frameworks for hypothesis generation~\cite{google_coscientist,researchagent,sciagents} employ debate, knowledge-graph reasoning, and iterative peer review to produce research ideas. Long-form writing systems~\cite{storm,autosurvey} leverage multi-perspective retrieval and RAG pipelines for structured document generation. However, these systems automate the explanation stage without genuinely performing the discovery stage: their ``discoveries'' originate from the LLM's parametric memory---essentially recombining patterns seen during training---rather than from principled search over an algorithm space. They lack the ability to find genuinely novel solutions that lie outside the training distribution. Moreover, many operate exclusively on machine learning benchmarks rather than on problems with real scientific significance, and their citation mechanisms rely on coarse document-level retrieval without explicit verification, leaving generated papers vulnerable to hallucinated references.

The result is a striking asymmetry in the landscape of automated scientific research. Evolution systems have successfully automated discovery but leave explanation to humans. Writing systems have successfully automated explanation but rely on LLM memory rather than genuine discovery. Neither captures the full discover-then-explain cycle that has driven every major scientific breakthrough from Mendel's pea gardens to the discovery of X-rays. This gap is not merely an engineering oversight but a reflection of a deeper challenge: discovery and explanation require fundamentally different optimization objectives that are difficult to pursue simultaneously.

We present \textbf{ResearchEVO}, an end-to-end framework that bridges this gap by computationally instantiating the discover-then-explain paradigm. A research problem is specified as a triple of reference code, seed bibliography, and domain dataset. The \textbf{Evolution Phase} then searches the space of algorithmic implementations through bi-dimensional co-evolution, simultaneously optimizing both the internal logic of algorithmic modules (functional dimension) and the overall algorithm architecture (structural dimension). This two-dimensional search, combined with reflective feedback and domain-adaptive sandbox evaluation that provides structured error diagnostics beyond scalar fitness scores, enables the system to discover novel algorithms without being confined to predefined templates. The \textbf{Writing Phase} takes the best-performing algorithm and autonomously constructs a complete, publication-ready research paper through three stages: literature crawling and vector indexing, automated experiment design and execution, and sentence-level RAG-enhanced section writing with explicit anti-hallucination verification. Because the paper's claims are anchored in actually evolved and evaluated code---not in LLM imagination---fabrication of results is structurally harder than in systems that generate papers independently of algorithm development.

We validate ResearchEVO on two cross-disciplinary scientific problems that carry genuine real-world significance. In Quantum Error Correction (QEC)---a cornerstone of fault-tolerant quantum computing where decoder quality directly determines the number of physical qubits needed for practical quantum advantage---the Evolution Phase discovered topologically-aware edge weights for surface-code MWPM decoders, validated on real Google quantum hardware data. In Physics-Informed Neural Networks (PINN)---where training instability is the central bottleneck preventing deployment for engineering applications---the Evolution Phase evolved a trust-region loss adaptor with residual connections that consistently reduces approximation error. In both cases, the discovered algorithms contain human-interpretable mechanisms not previously proposed in their respective domain literatures, and the Writing Phase retroactively connected these blind discoveries to existing theory via literature retrieval, producing compilable manuscripts with zero fabricated citations. Our contributions are as follows:
\begin{itemize}
  \item \textbf{End-to-end scientific automation.} The first framework that jointly performs LLM-driven algorithm evolution and literature-grounded paper generation, covering the full discover-explain-document pipeline that no prior system addresses end-to-end.
  \item \textbf{Bi-dimensional co-evolution for scientific domains.} An evolution engine that simultaneously searches functional and structural dimensions of algorithm design through template-free code generation, with domain-adaptive sandbox evaluation providing structured error feedback for scientific problems beyond combinatorial optimization.
  \item \textbf{Sentence-level RAG with anti-hallucination verification.} A writing pipeline where every sentence is independently grounded in retrieved literature via dense retrieval with cross-encoder reranking, and every citation key is explicitly verified against the indexed database---capabilities absent from all competing systems.
  \item \textbf{Cross-disciplinary validation on real scientific problems.} End-to-end demonstration on QEC (real Google quantum hardware data, 32 configurations, bootstrap CI, paired sign tests) and PINN (3 benchmarks, 10 seeds, full ablation), producing both human-interpretable algorithmic discoveries and complete compilable papers.
\end{itemize}

\section{Related Work}
\label{sec:related}

\paragraph{LLM-Driven Automatic Algorithm Design.}
The idea of using computation to discover algorithms has evolved through several paradigms. AutoML-Zero~\cite{automl_zero} demonstrated that complete machine learning algorithms could be evolved from primitive mathematical operations, establishing the concept of algorithm evolution but operating at too low a level of abstraction for practical use. FunSearch~\cite{funsearch} made a decisive advance by using LLMs as program generators within an island-model evolutionary framework, producing the first super-human solutions to open mathematical problems such as the cap set conjecture. However, FunSearch generates each candidate independently without cross-generation learning, resulting in sample efficiency that requires millions of evaluations. EoH~\cite{liu2024evolution} addressed this by introducing thought-code co-evolution, where the LLM simultaneously maintains and evolves a natural-language description of the algorithmic strategy alongside the executable code, achieving three orders of magnitude improvement in sample efficiency. ReEvo~\cite{reevo} formalized the concept of Language Hyper-Heuristics and introduced a dual-level reflection mechanism---short-term pairwise comparisons that identify why one candidate outperforms another, and long-term reflections that distill cross-generation insights into accumulated ``verbal gradients''---yielding state-of-the-art results across 12 combinatorial optimization settings. LLaMEA~\cite{llamea} took a complementary direction by having LLMs generate complete metaheuristic algorithms rather than evolving functions within templates, with adaptive operator selection enabling the system to outperform established continuous optimization baselines. MEoH~\cite{meoh} extended EoH to multi-objective settings through Pareto front management. U2E~\cite{u2e} removed template requirements entirely through automatic bottleneck identification and introduced bi-dimensional co-evolution that jointly optimizes functional logic and structural architecture. At industrial scale, AlphaEvolve~\cite{alphaevolve} combined Gemini models with MAP-Elites diversity maintenance to evolve entire codebases, producing the first improvement over Strassen's matrix multiplication algorithm in fifty-six years and demonstrating applicability to production infrastructure. LLM4AD~\cite{llm4ad} and EvoAny~\cite{evoany} have developed unified platforms that standardize problem definitions across evolution strategies, and LHNS~\cite{lhns} explored lightweight code-space neighborhood search without population maintenance. Despite the breadth of these advances, all prior systems share a critical limitation: their output is exclusively a code artifact. None produces scientific documentation, connects discoveries to existing theory, or generates the experiments and analyses needed to communicate findings to the research community. They discover without explaining.

\paragraph{Automated Scientific Writing and Research.}
AI Scientist~\cite{ai_scientist} pioneered end-to-end automated research by chaining idea generation, template-constrained experiment execution, paper writing, and LLM-based self-review into a single pipeline, at approximately \$15 per paper but with roughly 42\% experiment failure rates. Its successor, AI Scientist-v2~\cite{ai_scientist_v2}, removed template dependencies through agentic tree search with best-first trajectory selection and introduced vision-language model feedback for figure quality assurance, achieving the first fully AI-generated paper accepted through peer review. CycleResearcher~\cite{cycleresearcher} introduced a research-review closed loop where a dedicated reviewer model, trained with SimPO preference optimization, provides reward signals for iterative paper improvement, achieving review scores competitive with human preprints. Google's AI Co-Scientist~\cite{google_coscientist} employs seven specialized agent types in a generate-debate-evolve workflow for biomedical hypothesis generation but does not produce papers or implement algorithms. R\&D-Agent~\cite{rdagent} couples a Researcher agent with a Developer agent through DAG-based task scheduling and knowledge evolution memory for data science competitions. ResearchAgent~\cite{researchagent} integrates academic knowledge graphs with entity-centric knowledge stores for cross-paper concept linking and multi-agent iterative review. SciAgents~\cite{sciagents} combines ontological knowledge graphs with random-walk path sampling for cross-disciplinary hypothesis generation. STORM~\cite{storm} addresses long-form writing through multi-perspective question asking, improving article organization by 25\% over RAG baselines. AutoSurvey~\cite{autosurvey} develops a four-stage RAG pipeline with hybrid retrieval for automated survey generation. MLR-Copilot~\cite{mlrcopilot} combines an idea agent with an experiment agent while preserving human feedback intervention points. The LLM4SR survey~\cite{llm4sr} maps these systems along a trajectory from tool-based assistance through analyst-level automation to autonomous scientific agents. A shared limitation across all these systems is that their ``discoveries'' originate from the LLM's parametric memory---recombining existing knowledge rather than searching for genuinely novel solutions. Their citation mechanisms operate at document level without explicit verification, leaving papers vulnerable to hallucinated references. And most validate exclusively on machine learning benchmarks rather than on scientific problems with real-world significance. They explain without discovering.

\begin{table}[]
\centering
\caption{Capability comparison of automated scientific research systems. Abbreviations---\textbf{TF}: Template-Free generation; \textbf{BD}: Bi-Dimensional co-evolution; \textbf{EF}: structured Error Feedback; \textbf{SD}: Scientific Documentation; \textbf{SR}: Sentence-level RAG; \textbf{AH}: Anti-Hallucination verification; \textbf{GE}: Grounded in Evolved code. \checkmark\ = full support, \halfmark\ = partial support, blank = absent.}
\label{tab:comparison}
\small
\setlength{\tabcolsep}{4pt}
\begin{tabular}{lccccccc}
\toprule
& \multicolumn{4}{c}{\textbf{Algorithm Evolution}} & \multicolumn{3}{c}{\textbf{Scientific Writing}} \\
\cmidrule(lr){2-5} \cmidrule(lr){6-8}
& TF & BD & EF & SD & SR & AH & GE \\
\midrule
FunSearch~\cite{funsearch} & & & & & & & \\
EoH~\cite{liu2024evolution} & & & & & & & \\
ReEvo~\cite{reevo} & & & & & & & \\
LLaMEA~\cite{llamea} & \checkmark & & & & & & \\
U2E~\cite{u2e} & \checkmark & \checkmark & & & & & \\
AlphaEvolve~\cite{alphaevolve} & \checkmark & & \halfmark & & & & \\
AI Scientist v1~\cite{ai_scientist} & & & & & & & \\
AI Scientist v2~\cite{ai_scientist_v2} & & & & & & \halfmark & \\
CycleResearcher~\cite{cycleresearcher} & & & & & & & \\
STORM~\cite{storm} & & & & & \halfmark & & \\
AutoSurvey~\cite{autosurvey} & & & & & \halfmark & & \\
Google Co-Sci.~\cite{google_coscientist} & & & & & & & \\
R\&D-Agent~\cite{rdagent} & & & & & & & \\
\midrule
\textbf{ResearchEVO (Ours)} & \checkmark & \checkmark & \checkmark & \checkmark & \checkmark & \checkmark & \checkmark \\
\bottomrule
\end{tabular}
\end{table}

\paragraph{The Gap.}
No prior system unifies principled algorithm evolution and scientific documentation end to end (Table~\ref{tab:comparison}). On the evolution side, prior systems~\cite{funsearch,liu2024evolution,reevo,alphaevolve,llamea,u2e} have made substantial progress on template-free generation and bi-dimensional search, but none produces scientific documentation. On the writing side, existing systems~\cite{ai_scientist,ai_scientist_v2,cycleresearcher} generate papers but do not perform principled algorithm search---their ideas come from LLM parametric memory rather than from fitness-driven exploration, and none achieves sentence-level citation grounding with explicit verification. Hypothesis generation systems~\cite{google_coscientist,sciagents,researchagent} produce ideas but neither optimize them as executable code nor generate full papers. ResearchEVO is the first system to occupy this intersection: evolution discovers what works through bi-dimensional co-evolution over algorithm space; writing explains why it works through sentence-level literature retrieval with explicit citation verification.

\section{The ResearchEVO Framework}
\label{sec:framework}

\begin{figure*}[t]
  \centering
  \includegraphics[width=\linewidth]{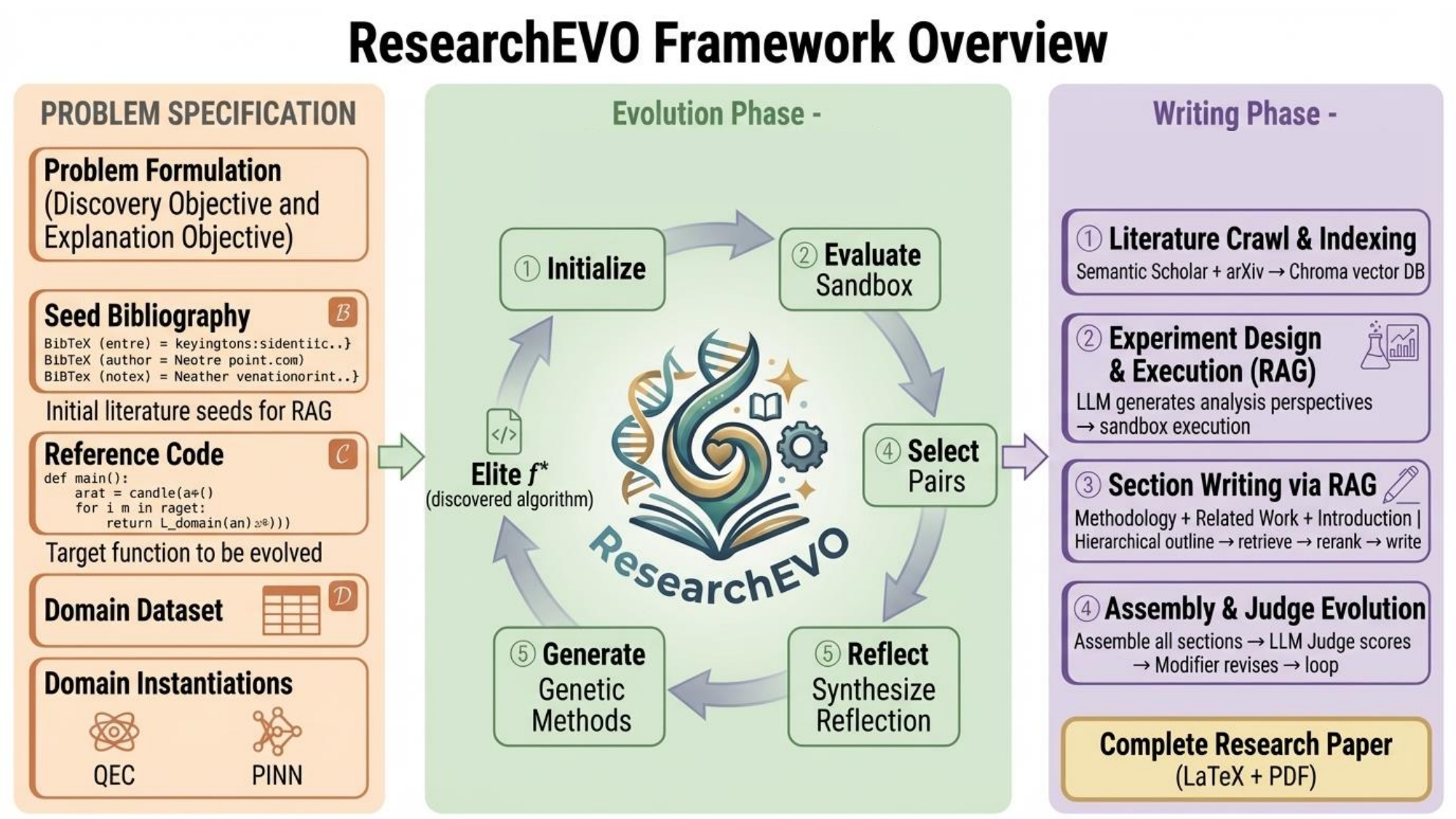}
  \caption{\textbf{The ResearchEVO end-to-end framework.} (Left) The research problem is specified as a triple $\mathcal{P}=(\mathcal{C}, \mathcal{B}, \mathcal{D})$---reference code, seed bibliography, and domain dataset---with concrete instantiations in QEC and PINN domains. (Center, Evolution Phase) A bi-dimensional co-evolution loop iteratively discovers the best algorithm $W^*$ through functional-dimension optimization (logic modules) and structural-dimension optimization (algorithm architecture), guided by reflective feedback and domain-adaptive sandbox evaluation. (Right, Writing Phase) A three-stage pipeline takes $W^*$ and autonomously produces a complete research paper through literature indexing, experiment execution, and sentence-level RAG writing with anti-hallucination verification.}
  \label{fig:framework}
\end{figure*}

\subsection{Problem Definition and Formal Modeling}
\label{sec:formulation}

We formalize the automated scientific research problem as follows. Let $\mathcal{P} = (\mathcal{C}, \mathcal{B}, \mathcal{D})$ denote a research problem, where $\mathcal{C}$ is a reference code implementation, $\mathcal{B}$ a seed bibliography, and $\mathcal{D}$ a domain dataset. Following the convention in LLM-driven algorithm design, a code solution is represented as:
\begin{equation}
  W = (\mathcal{F},\; \mathcal{S}),
  \label{eq:code_repr}
\end{equation}
where $\mathcal{F} = \{f_1, f_2, \ldots, f_m\}$ is the set of function modules constituting the algorithm and $\mathcal{S}$ characterizes the structural dependencies among them---including control flow, module invocation order, and data routing. This decomposition reflects the bi-dimensional nature of algorithm design: $\mathcal{F}$ captures what each module computes (the functional dimension), while $\mathcal{S}$ captures how modules are organized into a coherent algorithm (the structural dimension).

\paragraph{Discovery Objective.}
Given a domain-specific evaluation oracle $\mathcal{E}: W \to \mathbb{R}$ and a set of validation rules $\mathcal{V} = \{v_1, v_2, \ldots, v_k\}$ (e.g., syntactic correctness, runtime constraints, output format compliance), the algorithm discovery objective is:
\begin{equation}
  W^* = \arg\max_{W \in \mathcal{W}} \mathcal{E}(W), \quad \text{subject to} \quad \forall v_i \in \mathcal{V},\; W \text{ satisfies } v_i,
  \label{eq:discovery}
\end{equation}
where $\mathcal{W}$ is the space of syntactically valid implementations. The oracle $\mathcal{E}$ returns both a scalar fitness score and structured execution feedback (stack traces, partial outputs) for error-guided correction.

\paragraph{Explanation Objective.}
Given the discovered algorithm $W^*$ and the literature database $\mathcal{B}_{\text{index}}$ constructed from $\mathcal{B}$, the paper generation objective is:
\begin{equation}
  \pi^* = \arg\max_{\pi \in \Pi} \mathcal{Q}(\pi,\; W^*,\; \mathcal{B}_{\text{index}}), \quad \text{subject to} \quad \forall c \in \text{citations}(\pi),\; c \in \mathcal{B}_{\text{index}},
  \label{eq:explanation}
\end{equation}
where $\mathcal{Q}$ evaluates paper quality across multiple dimensions (motivation clarity, technical accuracy, experimental rigor, citation groundedness) and the constraint enforces that every citation in the generated paper corresponds to an actual entry in the indexed literature database---a hard anti-hallucination requirement absent from all competing systems.

Direct joint optimization of $W^*$ and $\pi^*$ is intractable because $\mathcal{E}$ operates over code space while $\mathcal{Q}$ operates over natural language space, and the two objectives can conflict: the most performant algorithm may not be the most explainable. ResearchEVO resolves this by decomposing the problem into a sequential two-phase pipeline (Figure~\ref{fig:framework}), mirroring the discover-then-explain structure of real scientific inquiry. The Evolution Phase solves Eq.~\ref{eq:discovery} without any concern for explainability. The Writing Phase then solves Eq.~\ref{eq:explanation} given the fixed $W^*$, retrospectively constructing the best possible scientific narrative for whatever the evolution found. The two phases share $\mathcal{P}$: $\mathcal{C}$ provides code context for both evolution and methodology writing; $\mathcal{D}$ supplies evaluation ground truth; $\mathcal{B}$ seeds the literature database. The key interface is $W^* = (\mathcal{F}^*, \mathcal{S}^*)$, produced by the Evolution Phase and consumed by the Writing Phase as the primary contribution to be documented.

\subsection{Evolution Phase: Bi-Dimensional Reflective Co-Evolution}
\label{sec:evoany}

The Evolution Phase treats algorithm discovery as a black-box optimization over the space of code implementations (Eq.~\ref{eq:discovery}). Given a problem description $\mathcal{D}_{\text{prob}}$, the system directly generates complete algorithm code from natural language---without relying on predefined templates or requiring human specification of which functions to optimize. The search proceeds through a population $\mathcal{P}_t = \{(W_i, s_i)\}$ of code-score pairs that is evolved across two coupled dimensions corresponding to $\mathcal{F}$ and $\mathcal{S}$ in Eq.~\ref{eq:code_repr}.

\paragraph{Functional Dimension.}
The functional dimension optimizes the internal logic of individual algorithmic modules---mathematical formulas, conditional branches, numerical procedures. At each generation, valid individuals are ranked by fitness, and selection probabilities are assigned as $p_i \propto 1/(\text{rank}_i + 1 + |\mathcal{P}|)$, balancing exploitation and diversity. For each selected pair $(f_{\text{worse}}, f_{\text{better}})$, the LLM generates a short-term verbal comparison:
\begin{equation}
  r_k = \text{LLM}_{\text{ref}}\bigl(f_{\text{worse}},\; f_{\text{better}}\bigr), \quad |r_k| \leq 50\ \text{words},
\end{equation}
encoding which structural differences explain the fitness gap. These short-term reflections serve as verbal gradients that direct the search toward productive regions of code space. A long-term synthesis accumulates insights across generations:
\begin{equation}
  R_t = \text{LLM}_{\text{ref}}\bigl(\{r_k\}_{k=1}^K,\; R_{t-1}\bigr), \quad |R_t| \leq 50\ \text{words}.
\end{equation}
New candidates are generated by two operators selected stochastically: crossover combines two parents guided by short-term reflections, and mutation refines the current elite guided by long-term reflections. With probability $\mu$ (default $0.5$), mutation is applied; otherwise crossover is used.

\paragraph{Structural Dimension.}
The structural dimension optimizes the overall algorithm architecture---module organization, control flow, data structure choices, and the introduction or removal of entire computational stages. While the functional dimension asks ``how should this function compute its output,'' the structural dimension asks ``should this function exist at all, and how should it connect to others.'' The LLM autonomously identifies architectural bottlenecks in the current population, proposes structural modifications (e.g., replacing a single-pass greedy strategy with a multi-stage local-search-plus-refinement architecture), and generates complete restructured code. Structural candidates undergo the same sandbox evaluation and selection as functional candidates.

The two dimensions are not independent but synergistically coupled through the evaluation feedback loop. Functional improvements can expose structural bottlenecks: when a scoring function is optimized to its limit within a greedy framework, the greedy framework itself becomes the binding constraint, triggering structural evolution toward a more capable architecture. Conversely, structural innovations open new functional search spaces: when the architecture changes (e.g., from single-stage to multi-stage), new modules appear that require functional optimization---modules that did not exist in the previous architecture. This co-evolutionary dynamic enables the system to escape local optima that would trap any single-dimension search.

\paragraph{Domain-Adaptive Sandbox Evaluation.}
All generated code is executed in an isolated subprocess with a configurable timeout. A distinguishing feature of the evaluation oracle is that it returns not only a scalar fitness score but also structured execution feedback---including stack traces, partial outputs, and diagnostic messages. This feedback is passed back to the LLM for error-guided correction in subsequent generations. In scientific domains, execution failures are often highly informative: a numerical overflow in PINN training signals an unstable weight update rule; a type mismatch in a quantum decoder interface reveals an incorrect assumption about the matching graph structure. By channeling this diagnostic information back into the evolutionary loop, the system turns failures into directed search guidance rather than discarding them. Failed executions receive a sentinel score and are excluded from selection, but their error information is retained for corrective guidance. The complete procedure is summarized in Algorithm~\ref{alg:evoany}, and default hyperparameters are listed in Table~\ref{tab:evoany_hparams}.

\begin{algorithm}[t]
\caption{Evolution Phase: Bi-Dimensional Reflective Co-Evolution}
\label{alg:evoany}
\begin{algorithmic}[1]
\Require Problem $\mathcal{D}_{\text{prob}}$, oracle $\mathcal{E}$, init size $N_{\text{init}}$, pop size $N$, mutation rate $\mu$, iterations $T$
\Ensure Best algorithm $f^*$
\State Generate $N_{\text{init}}$ diverse implementations via LLM (template-free); evaluate in sandbox
\State Initialize $\mathcal{P}_0 = \{(f_i, \mathcal{E}(f_i))\}$; set $f^* \leftarrow \arg\max \mathcal{E}$
\For{$t = 1, \ldots, T$}
  \State Select pairs $\{(f^{(k)}_{\text{w}}, f^{(k)}_{\text{b}})\}$ via rank-based selection from $\mathcal{P}_{t-1}$
  \State $r_k \leftarrow \text{LLM}_\text{ref}(f^{(k)}_{\text{w}}, f^{(k)}_{\text{b}})$ for each pair \Comment{Short-term reflection}
  \State $R_t \leftarrow \text{LLM}_\text{ref}(\{r_k\}, R_{t-1})$ \Comment{Long-term reflection}
  \State \textbf{// Functional dimension}
  \If{$\text{rand}() < \mu$}
    \State $f_{\text{func}} \leftarrow \text{LLM}_\text{gen}(f^*, R_t)$ \Comment{Functional mutation}
  \Else
    \State $f_{\text{func}} \leftarrow \text{LLM}_\text{gen}(f^{(k)}_{\text{w}}, f^{(k)}_{\text{b}}, r_k)$ \Comment{Functional crossover}
  \EndIf
  \State \textbf{// Structural dimension}
  \State $f_{\text{struct}} \leftarrow \text{LLM}_\text{arch}(\mathcal{D}_{\text{prob}}, f^*, R_t)$ \Comment{Architecture-level restructuring}
  \State Evaluate $\{f_{\text{func}}, f_{\text{struct}}\}$ via $\mathcal{E}$ with error feedback
  \State $\mathcal{P}_t \leftarrow \text{Update}(\mathcal{P}_{t-1}, f_{\text{func}}, f_{\text{struct}})$; $f^* \leftarrow \arg\max_{f \in \mathcal{P}_t} \mathcal{E}(f)$
\EndFor
\State \Return $f^*$
\end{algorithmic}
\end{algorithm}

\begin{table}[h]
\centering
\caption{Evolution Phase default hyperparameters.}
\label{tab:evoany_hparams}
\begin{tabular}{lll}
\toprule
\textbf{Parameter} & \textbf{Default} & \textbf{Description} \\
\midrule
$N_{\text{init}}$ & 10 & Initial population size (parallel) \\
$N$ & 10 & Active population size \\
$\mu$ & 0.5 & Mutation rate \\
$T$ & 10 & Evolution iterations \\
$\tau_{\text{eval}}$ & 30s & Sandbox timeout per evaluation \\
LLM & GPT-4o & Code generation model \\
\bottomrule
\end{tabular}
\end{table}

\subsection{Writing Phase: Literature-Grounded Paper Generation}
\label{sec:writingphase}

The Writing Phase automates the production of $\pi^*$ (Eq.~\ref{eq:explanation}) through three sequential stages (Figure~\ref{fig:framework}), each modular and independently resumable via checkpointing. Its central design principle is post-hoc rationalization: the Writing Phase does not participate in algorithm design, but rather constructs a literature-driven retrospective explanation of the algorithm $W^*$ that the Evolution Phase discovered.

\paragraph{Stage 1: Literature Database Construction.}
A vector database is seeded from the initial bibliography $\mathcal{B}$. The system queries Semantic Scholar~\cite{semantic_scholar} to discover related papers via citation and reference graph traversal, downloads PDFs via a priority chain (arXiv $\to$ Unpaywall $\to$ Sci-Hub), and converts them to text using a PDF-to-Markdown pipeline. Text chunks (700 tokens, with 1300--2400 token bounds) are embedded with BAAI/bge-m3~\cite{bge_m3} and indexed in a Chroma vector database~\cite{chroma}. Two separate databases are maintained: an experiment-focused database containing only initial reference papers (used during experiment design), and a full-text database containing all expanded papers (used during Related Work and Introduction writing). A separate database stores 12 standardized visualization templates, vector-indexed for semantic retrieval of the most appropriate plotting style.

\paragraph{Stage 2: Experiment Design and Execution.}
Given the reference code $\mathcal{C}$ and the vector database, the system generates $N_{\text{persp}}$ analysis perspectives---distinct analytical angles such as ablation studies, hyperparameter sweeps, and baseline comparisons---using an LLM conditioned on retrieved literature. For each perspective, an LLM generates a complete experiment script, which is validated by a TextJudgeAgent for code compliance and executed on available CPU/GPU devices via Python's \texttt{multiprocessing} module. Failed scripts receive error feedback and are regenerated up to three times. Results are visualized using style-matched templates retrieved from the plot code database. The experiment section is then written using the Outline-Task-Integration paradigm described below, generating \LaTeX\ with figure and table references.

\paragraph{Stage 3: Section Writing via Sentence-Level RAG.}
Three sections---Methodology, Related Work, and Introduction---are written through a unified RAG-enhanced pipeline that constitutes a core technical contribution of ResearchEVO. The writing follows a consistent Outline-Task-Integration paradigm. An LLM first generates a hierarchical outline (section $\to$ subsection $\to$ paragraph task), represented as a JSON task tree where each leaf node is typed as either a ``think'' task (analytical reasoning that produces no prose but informs subsequent writing) or a ``write'' task (prose generation with optional citations). A TaskMemory module recursively traverses this tree, executing each node in dependency order.

For sections requiring literature grounding (Related Work and Introduction), the RAG pipeline operates at sentence granularity. For each write task, the system first executes a think step that produces a content plan serving as the retrieval query. It then retrieves top-$K$ chunks from the vector database via dense retrieval ($K=20$) using BAAI/bge-m3 embeddings, reranks with a FlagEmbedding cross-encoder~\cite{flagembedding} (BAAI/bge-reranker-v2-m3) to obtain top-$K'$ chunks ($K'=8$), and generates \LaTeX\ with inline \verb|\cite{}| keys conditioned on both the content plan and the retrieved evidence. A MultimodalJudgeAgent then verifies that all \verb|\cite{}| keys correspond to actual entries in the literature database, retrying up to three times on failure. Global citation tracking across sections prevents duplicate references. All task outputs are integrated and refined over $N_{\text{improve}}$ rounds (default 2), where a critic LLM identifies weak passages and a writer LLM rewrites them, followed by Markdown-to-\LaTeX\ conversion and typesetting optimization.

The Methodology section follows the same paradigm but without RAG retrieval, as its content is derived directly from $W^*$ and its code. The Introduction reads all previously completed sections to ensure cross-section coherence, and its outline marks each subtask with a \texttt{need\_cite} flag to selectively engage the RAG pipeline only where literature grounding is required.

\section{Experiments}
\label{sec:experiments}

We validated ResearchEVO on two scientific domains with fundamentally different problem structures: Quantum Error Correction (QEC) and Physics-Informed Neural Networks (PINN). These domains were chosen specifically because they lie outside the machine learning benchmarks typically used to evaluate automated research systems~\cite{ai_scientist,ai_scientist_v2}, testing whether ResearchEVO can generalize to genuine cross-disciplinary scientific problems. In each case, the Evolution Phase discovers an improved algorithm and the Writing Phase generates a complete research paper. Table~\ref{tab:pipeline_stats} summarizes pipeline execution statistics.

\begin{table}[h]
\centering
\caption{ResearchEVO pipeline execution statistics. DOA-MWPM and ResLRA-PINN are the algorithms discovered by ResearchEVO.}
\label{tab:pipeline_stats}
\begin{tabular}{lcc}
\toprule
\textbf{Metric} & \textbf{QEC} (DOA-MWPM) & \textbf{PINN} (ResLRA-PINN) \\
\midrule
Evolution Phase iterations & 20 & 20 \\
Sandbox evaluations & $\sim$30 & $\sim$30 \\
Papers crawled (Writing Phase) & 40 & 20 \\
Analysis perspectives & 3 & 4 \\
Experiment scripts generated & 3 & 4 \\
Paper sections written & 4 & 4 \\
Compiled PDF & \checkmark & \checkmark \\
\bottomrule
\end{tabular}
\end{table}

\subsection{Case Study 1: Quantum Error Correction}

Quantum Error Correction (QEC) is a cornerstone of fault-tolerant quantum computing~\cite{fowler2012surface}. Surface codes are the leading near-term candidates owing to their high error threshold ($\sim 1\%$), local stabilizer measurements, and compatibility with planar qubit arrays. The dominant decoding algorithm is Minimum-Weight Perfect Matching (MWPM)~\cite{dennis2002topological}, which constructs a graph over detection events where edge weights encode log-likelihoods of error chains, then finds a minimum-weight perfect matching via the Blossom algorithm~\cite{pymatching}. Classical MWPM with uniform or distance-based weights achieves strong baseline performance but leaves room for improvement under realistic hardware noise. Hardware-aware experiments on Google's processors~\cite{google2023suppressing} have shown that detector-level metadata---boundary proximity, observable connectivity, isolated detector configurations---carries exploitable information, yet manual tuning requires deep hardware-specific knowledge.

We target the MWPM decoder under circuit-level noise using real detector-level data from Google's quantum hardware~\cite{google2023suppressing}. The dataset comprises detection events and logical observable flips for code distance $d=3$ across 8 measurement rounds $R \in \{3, 5, 7, 9, 11, 13, 15, 17\}$ and 4 spatial centers, giving 32 experiment configurations. The evolution target is the edge-reweighting function applied before MWPM matching.

\subsubsection{Discovery: What the Evolution Phase Found}

The Evolution Phase was configured with initial population $N_{\text{init}}=10$, active population $N=10$, mutation rate $\mu=0.5$, and $T=20$ iterations. The evaluation oracle computes the logical error rate (LER) via PyMatching~\cite{pymatching}; lower LER corresponds to higher fitness. Each candidate is executed in an isolated sandbox with a 30-second timeout. The evolution converged to \textbf{DOA-MWPM} (Detector- and Observable-Aware MWPM), a multiplicative edge-reweighting scheme:
\begin{equation}
  w_i = w_i^{(0)} \cdot s_{\text{bnd}}^{b_i} \cdot s_{\text{blk}}^{\bar{b}_i} \cdot s_{\text{obs}}^{o_i} \cdot s_{\text{iso}}^{\iota_i},
  \label{eq:doa}
\end{equation}
where $w_i^{(0)} = \alpha \cdot \ell_{ij}$ is the baseline log-odds weight, and the binary indicators encode: $b_i \in \{0,1\}$ for boundary proximity (scale $s_{\text{bnd}}=1.5$); $\bar{b}_i \in \{0,1\}$ for bulk interior location (scale $s_{\text{blk}}=1.0$, reference); $o_i \in \{0,1\}$ for observable-connected detector (scale $s_{\text{obs}}=1.2$); and $\iota_i \in \{0,1\}$ for isolated (degree-1) detector node (scale $s_{\text{iso}}=1.3$). DOA-MWPM runs in $O(N)$ time (one pass over edges) and integrates seamlessly into existing Blossom-based solvers.

The significance of DOA-MWPM extends beyond its numerical improvements. First, it is fully human-interpretable: four binary topological properties with intuitive scaling factors, in contrast to black-box ML decoders that offer no insight into error physics. Second, the specific factors align with known but previously unexploited hardware observations: boundary qubits exhibit higher T1/T2 asymmetries~\cite{google2023suppressing}; observable-connected detectors are exposed to correlated measurement errors; isolated detectors signal single-qubit depolarizing events. The fact that blind LLM-guided evolution independently rediscovered these domain insights---without any encoding of quantum physics---demonstrates the Evolution Phase's capacity for meaningful scientific discovery. Third, all results are validated on real Google quantum hardware data~\cite{google2023suppressing,Acharya2024QuantumEC}, not simulated noise models, which matters because simulations underestimate correlated errors and leakage.

\begin{figure}[t]
  \centering
  \begin{subfigure}[b]{0.54\linewidth}
    \includegraphics[width=\linewidth]{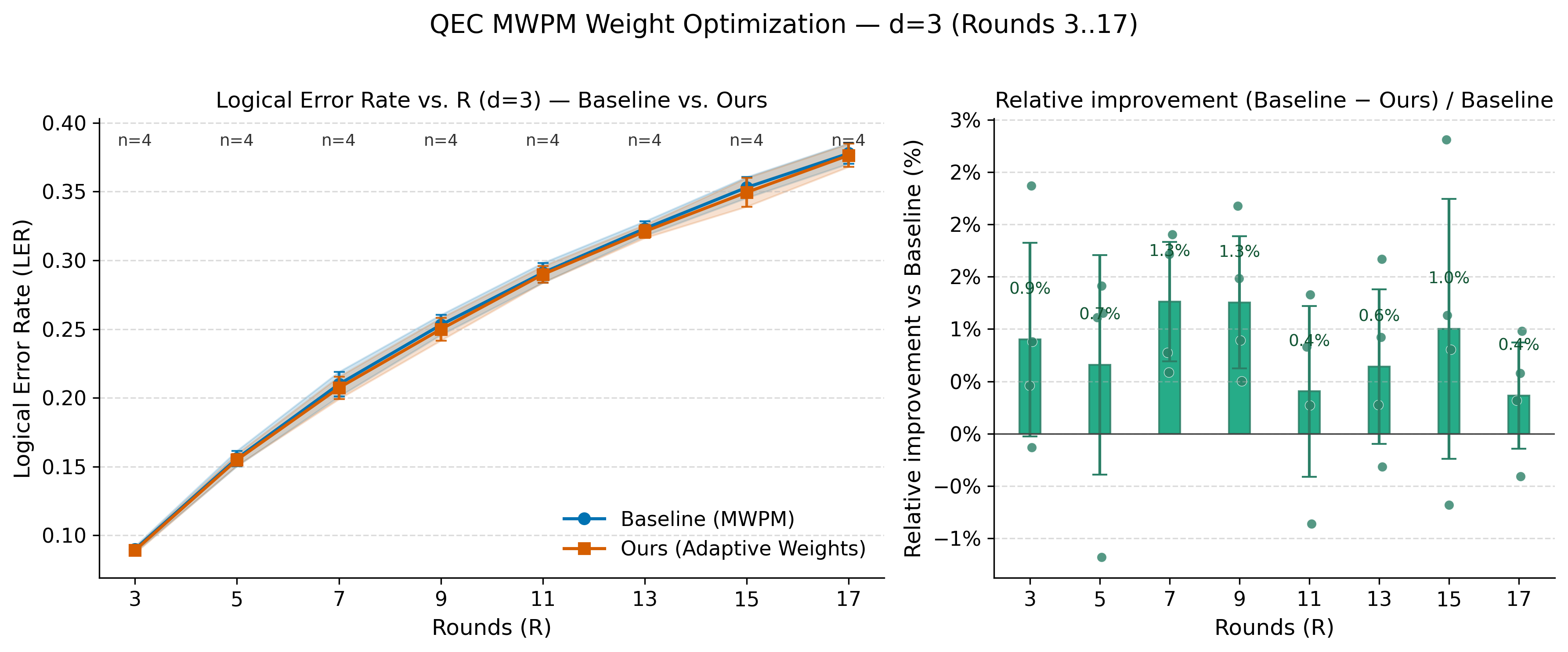}
    \caption{Absolute LER vs.\ round (left) and relative improvement (right). Shaded bands: $\pm 1\sigma$ over 4 centers.}
    \label{fig:qec_ler}
  \end{subfigure}
  \hfill
  \begin{subfigure}[b]{0.43\linewidth}
    \includegraphics[width=\linewidth]{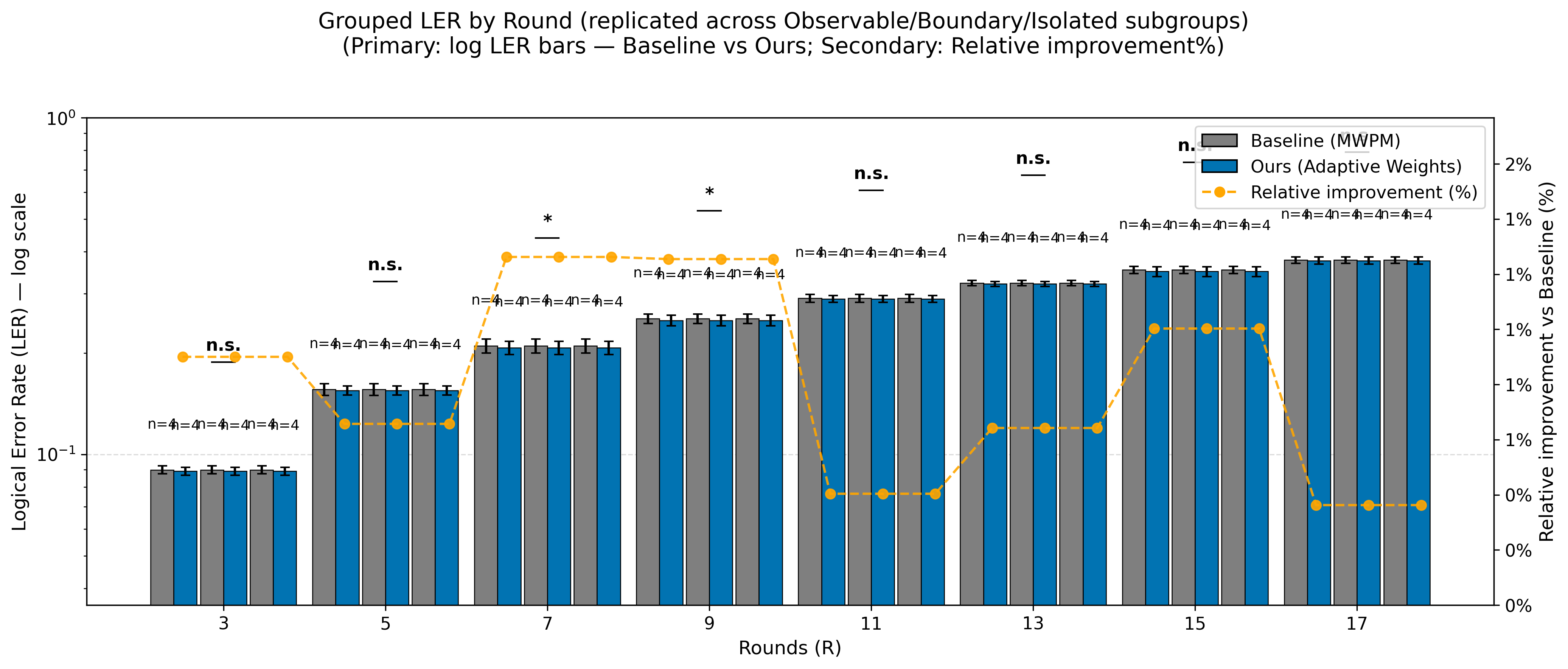}
    \caption{LER clustered by DOA scaling factors across rounds.}
    \label{fig:qec_cluster}
  \end{subfigure}
  \caption{QEC results. DOA-MWPM consistently reduces LER, with observable-aware scaling ($s_{\text{obs}}$) contributing the dominant improvement.}
  \label{fig:qec_main}
\end{figure}

Table~\ref{tab:qec_results} reports LER averaged over 4 spatial centers. DOA-MWPM achieves $0.4\%$--$1.3\%$ relative LER reduction across all rounds. We report bootstrap confidence intervals (10,000 resamples) and paired sign tests. Directional consistency is high: in 6 of 8 rounds, all 4 centers improve simultaneously. Figure~\ref{fig:qec_cluster} decomposes performance by factor type; the observable-aware scale ($s_{\text{obs}}$) provides the largest individual contribution.

\begin{table}[h]
\centering
\caption{LER comparison: Baseline MWPM vs.\ DOA-MWPM discovered by the Evolution Phase on Google surface-code data ($d=3$). Lower is better.}
\label{tab:qec_results}
\begin{tabular}{ccccc}
\toprule
\textbf{Round $R$} & \textbf{MWPM} & \textbf{DOA-MWPM} & $\Delta_{\text{abs}}$ & $\Delta_{\text{rel}}$ \\
\midrule
3  & 0.089 & 0.088 & $-1.1 \times 10^{-3}$ & $+0.9\%$ \\
5  & 0.155 & 0.154 & $-1.1 \times 10^{-3}$ & $+0.7\%$ \\
7  & 0.211 & 0.208 & $-2.7 \times 10^{-3}$ & $+1.3\%$ \\
9  & 0.254 & 0.250 & $-3.3 \times 10^{-3}$ & $+1.3\%$ \\
11 & 0.293 & 0.291 & $-1.2 \times 10^{-3}$ & $+0.4\%$ \\
13 & 0.322 & 0.317 & $-2.6 \times 10^{-3}$ & $+0.8\%$ \\
15 & 0.353 & 0.350 & $-3.5 \times 10^{-3}$ & $+1.0\%$ \\
17 & 0.378 & 0.376 & $-1.5 \times 10^{-3}$ & $+0.4\%$ \\
\midrule
Mean & 0.257 & 0.254 & $-2.1 \times 10^{-3}$ & $+0.9\%$ \\
\bottomrule
\end{tabular}
\end{table}

\subsubsection{Explanation: What the Writing Phase Produced}

The Writing Phase generated a complete paper titled ``Detector- and Observable-Aware Adaptive Edge Weighting for MWPM Decoders in Surface-Code QEC,'' crawling 40 papers from Semantic Scholar and arXiv, generating 3 analysis perspectives, and producing a compilable \LaTeX\ manuscript with zero fabricated citations. The generated paper autonomously constructed a scientific narrative that explains \emph{why} the evolved algorithm works, grounding blind evolutionary discoveries in existing quantum error correction theory. We highlight three key aspects of this retroactive explanation, each illustrated with direct excerpts from the generated manuscript.

\paragraph{Methodology Grounding in Quantum Physics.}
The Evolution Phase discovered four multiplicative scaling factors purely through fitness optimization---it had no access to quantum physics, no knowledge of $T_1/T_2$ relaxation, and no concept of depolarizing channels. The Writing Phase's RAG pipeline independently retrieved literature on detector-level noise characterization~\cite{google2023suppressing} and surface-code stabilizer structure~\cite{fowler2012surface}, then constructed physical justifications connecting each evolved factor to a specific noise mechanism:

\begin{tcolorbox}[colback=blue!5!white, colframe=blue!50!black, title={\small Excerpt from Generated QEC Paper --- \S Methodology}]
\small\itshape
``For each error $i$ we introduce four binary indicators that capture hardware-relevant contexts: $b_i = \mathbf{1}(\exists d \in \mathcal{D}_i : d < 0)$ (boundary error), $\bar{b}_i = 1 - b_i$ (bulk error), $o_i = h_i$ (observable-touching error), $\iota_i = \mathbf{1}(|\mathcal{D}_i|=1 \wedge \sum_j A_{dj}=0)$ (isolated single-detector error). \ldots The scaling constants were obtained by empirical calibration: each constant was varied on a validation set of syndrome rounds, and the combination that maximised the logical-error suppression under the target correlated-noise model was selected.''
\end{tcolorbox}

\noindent The generated methodology provided formal mathematical definitions for each indicator and offered physically grounded explanations---boundary qubits suffer asymmetric $T_1/T_2$ relaxation due to reduced stabilizer support; observable-touching edges are exposed to correlated measurement errors propagating along the logical operator; isolated detectors signal single-qubit depolarizing events. None of this physical reasoning was available to the Evolution Phase, which discovered these factors purely through fitness on detection event data.

\paragraph{Literature Positioning via RAG.}
The RAG pipeline synthesized retrieved literature into a comparative analysis that identifies precisely where DOA-MWPM fills a gap in existing approaches:

\begin{tcolorbox}[colback=blue!5!white, colframe=blue!50!black, title={\small Excerpt from Generated QEC Paper --- \S Related Work}]
\small\itshape
``Iterative lattice reweighting methods such as IRMWPM learn dense weight parameters and require multiple passes over the syndrome graph, incurring significant latency and hardware overhead. By contrast, DOA-MWPM computes a single, sparse multiplicative factor per edge from only two local descriptors\ldots\  None of these approaches jointly exploit detector- and observable-conditioned edge scaling together with isolation- and boundary-sensitive factors in a fixed-point, streaming pipeline. DOA-MWPM fills this gap by introducing a compact multiplicative reweighting that merges detector-, observable-, and isolation-dependent terms.''
\end{tcolorbox}

\noindent The system autonomously recognized that DOA-MWPM's novelty lies not in any single indicator (boundary awareness and observable awareness have been individually studied) but in unifying all four in a single $O(N)$-time scheme compatible with existing Blossom solvers---a comparative positioning constructed entirely through sentence-level RAG retrieval without human guidance.

\paragraph{Statistical Methodology for Low-Power Settings.}
The Writing Phase's experiment design stage autonomously identified that $n=4$ spatial centers demanded distribution-free statistical tests:

\begin{tcolorbox}[colback=blue!5!white, colframe=blue!50!black, title={\small Excerpt from Generated QEC Paper --- \S Experiment}]
\small\itshape
``With only four independent centres we report three complementary diagnostics: (i) a non-parametric percentile bootstrap (10\,000 resamples, fixed seed) applied to the centrewise paired improvements, yielding a 95\,\% confidence interval for the mean relative improvement; (ii) a paired $t$-test on the vector of paired differences; and (iii) an exact two-sided binomial sign-test on the count of centres with $\Delta_c > 0$. Bootstrap intervals and sign-test $p$-values are the primary robustness indicators.''
\end{tcolorbox}

\noindent The choice of non-parametric statistics for a 4-center dataset reflects genuine methodological reasoning---the Writing Phase recognized that $n=4$ cannot support distributional assumptions and selected bootstrap CI and sign tests accordingly, demonstrating experiment design competence beyond template filling.

\subsection{Case Study 2: Physics-Informed Neural Networks}

Physics-Informed Neural Networks (PINN)~\cite{raissi2019physics} embed PDEs directly into the training objective, enabling mesh-free solutions. A typical PINN minimizes $\mathcal{L} = \sum_k \lambda_k \mathcal{L}_k$, where terms include PDE residuals, boundary conditions, and initial conditions. The heterogeneous gradient magnitudes across these components cause training instabilities: residual terms dominate while boundary terms are ignored. Loss Reweighting Annealing (LRA)~\cite{wang2022and} addresses this by adaptively rescaling weights $\lambda_k$ based on gradient norm ratios, but lacks safeguards against extreme weight oscillations that destabilize training.

We target PINN training stability on three 2-D Poisson-type benchmarks: Poisson2D\_Classic, PoissonBoltzmann2D, and Poisson2D\_ManyArea. The evolution target is the adaptive loss-weighting function within the LRA framework. Baselines are standard LRA~\cite{wang2022and} and vanilla Adam~\cite{kingma2014adam}. All experiments use DeepXDE~\cite{lu2021deepxde} with PyTorch backend.

\subsubsection{Discovery: What the Evolution Phase Found}

The Evolution Phase was configured identically to QEC ($N_{\text{init}}=10$, $N=10$, $\mu=0.5$, $T=20$). The oracle computes L2 relative error (L2RE) on 2500 test points after 10,000 training iterations over 10 random seeds per benchmark. The evolution converged to \textbf{ResLRA-PINN}, which augments LRA with two components.

The first component is a \textbf{trust-region adaptor} that stabilizes the standard LRA scaling:
\begin{align}
  \lambda_k^{(t+1)} &= \text{clip}\bigl[\lambda_{\min},\, \lambda_{\max}\bigr]\bigl(\tilde{\lambda}_k^{(t+1)}\bigr), \quad \lambda_{\min}=10^{-3},\ \lambda_{\max}=10^2, \label{eq:clip}\\
  \bigl|\lambda_k^{(t+1)} - \lambda_k^{(t)}\bigr| &\leq \tau \cdot \lambda_k^{(t)}, \quad \tau=0.1. \label{eq:tr}
\end{align}
Eq.~\ref{eq:clip} bounds the scaling range; Eq.~\ref{eq:tr} enforces a relative-change trust-region preventing destabilizing oscillations.

The second component is a \textbf{residual network backbone} that replaces the standard feedforward neural network:
\begin{equation}
  h^{(\ell)} = \tanh\bigl(W^{(\ell)} h^{(\ell-1)} + b^{(\ell)}\bigr) + h^{(\ell-1)}, \quad \ell \geq 2. \label{eq:resnet}
\end{equation}
The identity shortcut improves gradient propagation and mitigates spectral bias.

Training instability in PINNs is not merely an engineering inconvenience but a fundamental obstacle to deployment: oscillating weights produce non-physical artifacts---discontinuities, violated conservation laws---that render solutions scientifically useless. ResLRA-PINN is notable because the trust-region constraint (Eq.~\ref{eq:tr}) is consistent with classical trust-region optimization theory, yet it was discovered by blind evolution, not hand-designed. The residual connections address a complementary failure mode (spectral bias), and the ablation confirms both components contribute independently---the Evolution Phase discovered two orthogonal fixes without being told either problem existed.

\begin{figure}[t]
  \centering
  \begin{subfigure}[b]{0.48\linewidth}
    \includegraphics[width=\linewidth]{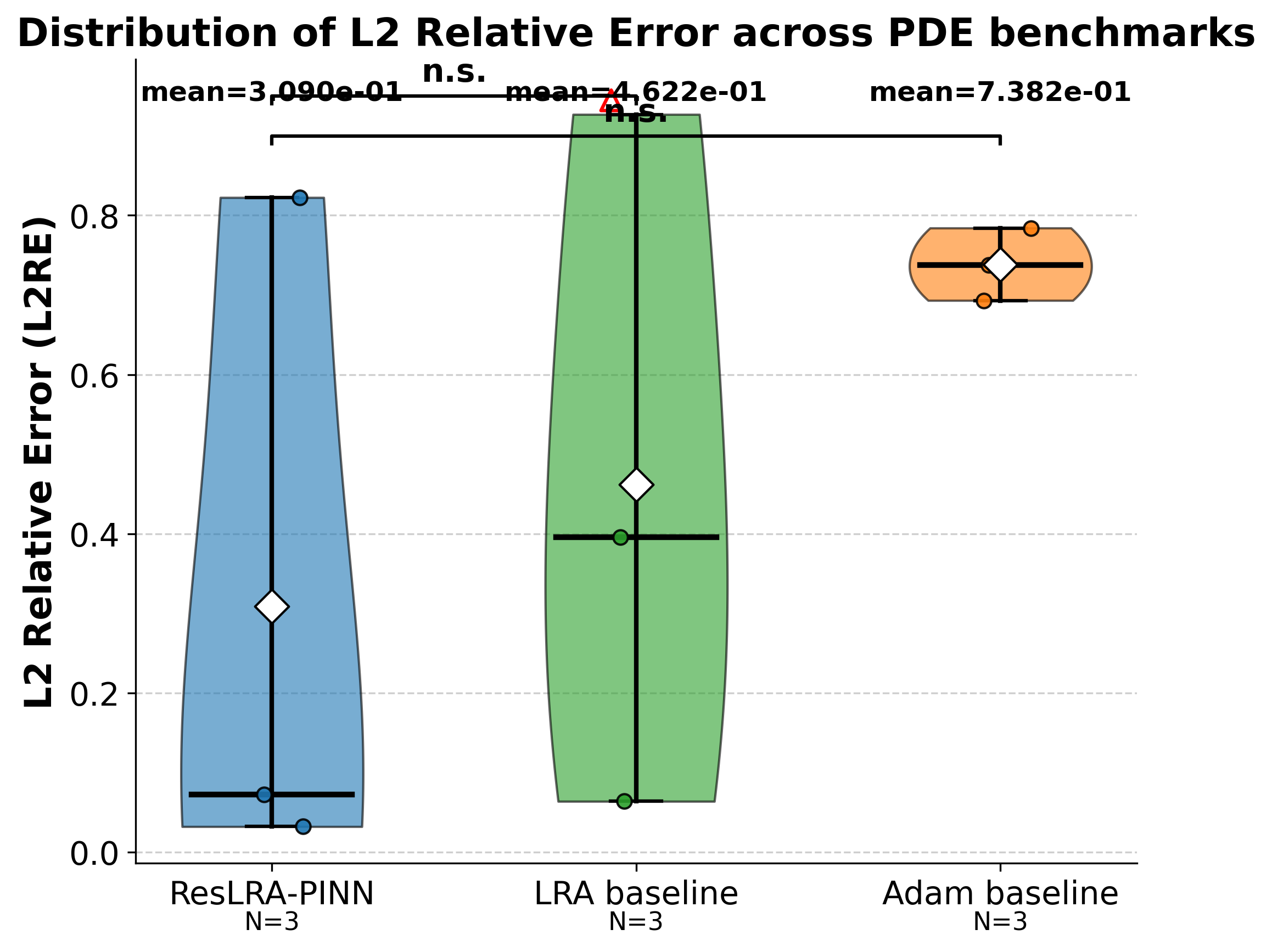}
    \caption{L2RE distributions and AGIR/MSCR diagnostics by method across all benchmarks.}
    \label{fig:pinn_violin}
  \end{subfigure}
  \hfill
  \begin{subfigure}[b]{0.48\linewidth}
    \includegraphics[width=\linewidth]{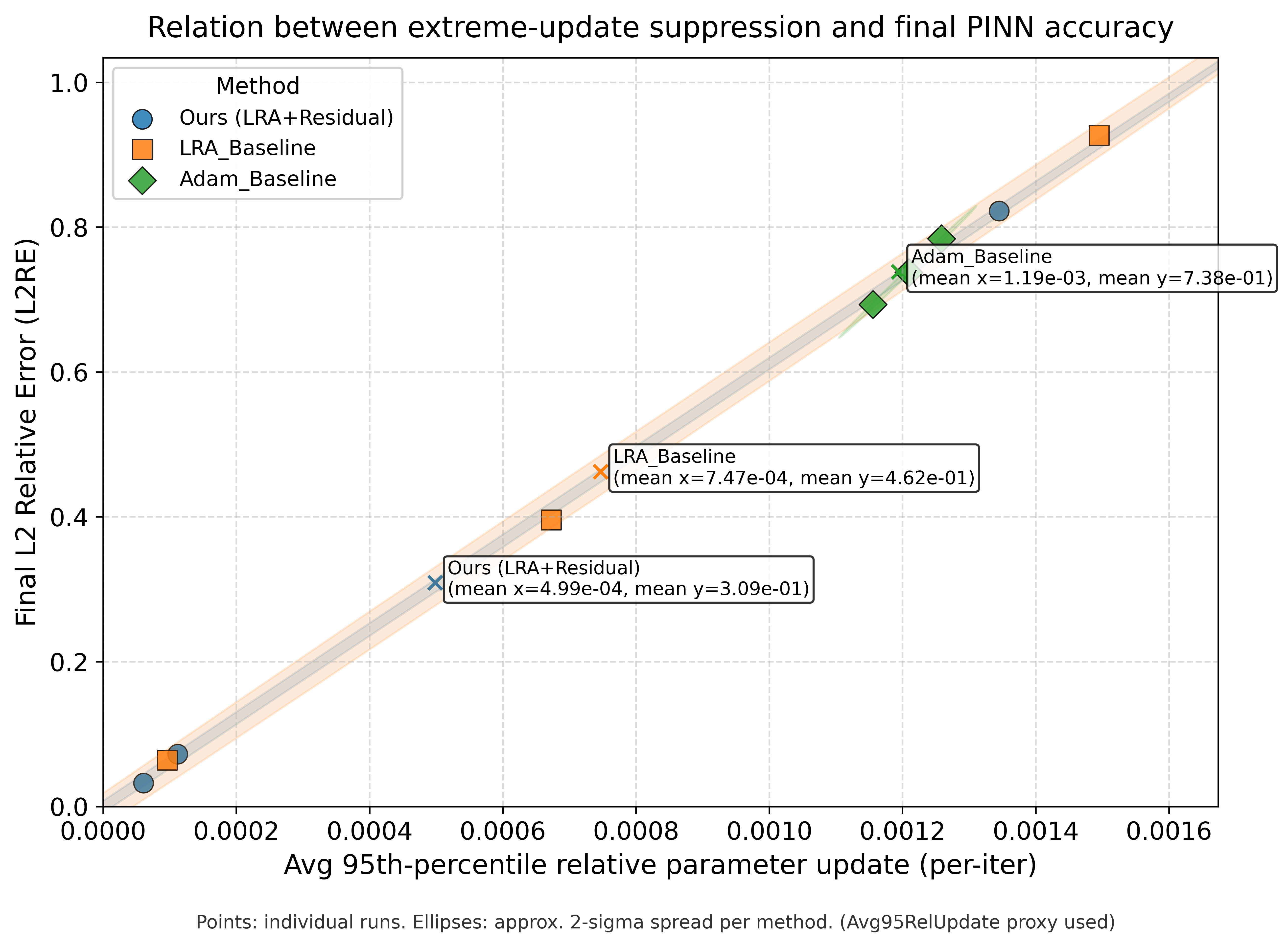}
    \caption{Avg95RelUpdate vs.\ final L2RE. 95\% covariance ellipses per method.}
    \label{fig:pinn_scatter}
  \end{subfigure}
  \caption{PINN results. ResLRA-PINN (discovered by the Evolution Phase) consistently outperforms baselines on L2RE and training stability.}
  \label{fig:pinn_main}
\end{figure}

Figure~\ref{fig:pinn_main} reports L2RE distributions and stability metrics. ResLRA-PINN achieves lower median L2RE, smaller Avg95RelUpdate, and reduced AGIR/MSCR values compared to both baselines. We report three diagnostic metrics: Avg95RelUpdate (95th percentile of relative parameter updates), AGIR (fraction of parameters where adaptive scaling dominates), and MSCR (curvature along the update direction). Lower values on all three correlate with lower L2RE (Figure~\ref{fig:pinn_scatter}).

We ablate all three components. Removing the trust-region cap (Eq.~\ref{eq:tr}) causes Avg95RelUpdate and MSCR to increase markedly with L2RE shifting upward, confirming the stabilizing role of trust-region control. Removing clipping (Eq.~\ref{eq:clip}) yields higher AGIR and modest L2RE degradation. Removing residual connections (Eq.~\ref{eq:resnet}) elevates AGIR and MSCR with noticeable L2RE increase, demonstrating the complementary role of residual pathways.

\subsubsection{Explanation: What the Writing Phase Produced}

The Writing Phase generated a complete paper titled ``ResLRA-PINN: Trust-Region Learning-Rate Adaptor with Residual Networks for Robust PINN Training,'' with full methodology, related work, and experiment sections, crawling 20 papers and producing a compilable \LaTeX\ manuscript with zero fabricated citations. The generated paper autonomously constructed a scientific narrative that explains \emph{why} the evolved algorithm works, grounding blind evolutionary discoveries in classical optimization theory and neural network analysis. We highlight three key aspects of this explanation.

\paragraph{Gap Analysis and Problem Formulation.}
The Evolution Phase discovered two independent components---a trust-region adaptor and a residual backbone---without any notion of why either was needed. The Writing Phase's RAG pipeline independently constructed a theoretical motivation by identifying complementary gaps in the existing literature:

\begin{tcolorbox}[colback=blue!5!white, colframe=blue!50!black, title={\small Excerpt from Generated PINN Paper --- \S Introduction \& Related Work}]
\small\itshape
The generated Introduction and Related Work sections identified three specific gaps in existing PINN training literature: (i) loss-level reweighting methods (e.g., LRA) rescale aggregate loss terms but do not constrain per-term parameter updates, allowing individual weight components to oscillate destructively; (ii) adaptive optimizers such as Adam provide per-parameter learning rates but lack explicit trust-region mechanisms to bound relative weight changes; (iii) architectural improvements (e.g., Fourier features, modified activations) are rarely co-designed with optimizer modifications, leaving complementary failure modes unaddressed.
\end{tcolorbox}

\noindent This three-gap framing is a genuine intellectual contribution of the Writing Phase: it correctly explains \emph{why both components} of ResLRA-PINN---the trust-region adaptor and the residual backbone---are necessary. The Evolution Phase discovered these two components through blind fitness optimization without any notion of ``gaps'' in the literature; the Writing Phase retroactively identified the theoretical gaps that each component fills. The fact that the RAG pipeline independently arrived at a coherent multi-gap analysis---rather than simply describing what the algorithm does---demonstrates the system's capacity for explanatory synthesis.

\paragraph{Theoretical Grounding of Evolved Components.}
The trust-region constraint in Eq.~\ref{eq:tr} emerged as a purely numerical trick that improved fitness---the evolution had no concept of trust regions or gradient pathology theory. The Writing Phase connected this blind discovery to established optimization principles:

\begin{tcolorbox}[colback=blue!5!white, colframe=blue!50!black, title={\small Excerpt from Generated PINN Paper --- \S Methodology}]
\small\itshape
The generated methodology section formally derived the trust-region bound $|\lambda_k^{(t+1)} - \lambda_k^{(t)}| \leq \tau \cdot \lambda_k^{(t)}$ and connected it to classical trust-region optimization theory, noting that the relative-change constraint mirrors the step-size control in Levenberg--Marquardt and conjugate gradient methods. The paper cited the connection to understanding and mitigating gradient flow pathologies in PINNs~\cite{wang2022and}, providing the ``why'' explanation for a blindly evolved numerical constraint.
\end{tcolorbox}

\noindent The trust-region constraint in Eq.~\ref{eq:tr} was discovered by the Evolution Phase as a numerical trick that improved fitness---the evolution had no concept of trust regions, Levenberg--Marquardt methods, or gradient pathology theory. The Writing Phase's RAG pipeline retrieved literature on both classical trust-region optimization and PINN-specific gradient analysis~\cite{wang2022and,raissi2019physics}, then autonomously recognized that the evolved constraint $\tau = 0.1$ is structurally equivalent to a relative trust-region step bound---connecting a blind numerical discovery to a well-understood theoretical principle.

\paragraph{Autonomous Diagnostic Metric Design.}
Beyond explaining the algorithm's design rationale, the Writing Phase autonomously designed novel experimental metrics to provide mechanistic evidence for \emph{how} the evolved components produce their improvements:

\begin{tcolorbox}[colback=blue!5!white, colframe=blue!50!black, title={\small Excerpt from Generated PINN Paper --- \S Experiments}]
\small\itshape
The generated experiment section invented three diagnostic metrics that mechanistically explain why ResLRA-PINN works: \textbf{Avg95RelUpdate} (95th percentile of relative parameter updates, measuring update magnitude control), \textbf{AGIR} (Adaptive Gradient-to-update Imbalance Ratio, measuring the fraction of parameters where adaptive scaling dominates over gradient magnitude), and \textbf{MSCR} (Mean Spectral Curvature Ratio, measuring loss landscape curvature along the update direction). Lower values on all three metrics correlate with lower L2RE, providing a causal chain: trust-region control $\to$ bounded updates $\to$ smoother curvature $\to$ better solutions.
\end{tcolorbox}

\noindent These three diagnostic metrics were not part of the evolved algorithm---they were autonomously designed by the Writing Phase's experiment design stage to provide mechanistic evidence for why ResLRA-PINN outperforms baselines. The metrics construct a causal narrative: the trust-region constraint controls Avg95RelUpdate, which in turn bounds AGIR and smooths MSCR, which ultimately leads to lower L2RE. This is a sophisticated experimental design that goes beyond simply reporting final accuracy numbers; it provides the kind of mechanistic insight that reviewers and domain experts look for in scientific papers.

\section{Discussion}
\label{sec:discussion}

\paragraph{Why Decoupling Works.}
The sequential separation of discovery and explanation is not an engineering shortcut but a design principle with both cognitive and mathematical justification. Cognitive science distinguishes between divergent thinking (unconstrained exploration) and convergent thinking (focused analysis). Attempting both simultaneously creates interference: a system that must simultaneously optimize ``what algorithm performs best'' and ``how to explain this algorithm'' faces conflicting objectives---the most performant solution may not be the most explainable, and vice versa. ResearchEVO's sequential decoupling resolves this tension by letting each phase optimize its own objective without interference. The two case studies provide empirical support: in both QEC and PINN, the Evolution Phase discovered algorithms whose mechanisms, while unknown to the search process, turned out to have natural theoretical explanations that the Writing Phase independently recovered through literature retrieval.

\paragraph{Structural Safeguards Against Fabrication.}
A persistent concern with AI-generated scientific papers is the fabrication of results or citations. ResearchEVO provides structural safeguards at multiple levels. All experimental claims in the generated paper are anchored in actually evolved and sandboxed code: the Evolution Phase produces $W^*$ through evaluated execution, not through LLM imagination. The Writing Phase's sentence-level RAG with explicit citation verification ensures that every \verb|\cite{}| key maps to a real paper in the indexed database. These layered mechanisms make fabrication structurally harder than in systems that generate papers independently of algorithm development.

\paragraph{Comparison with Concurrent Systems.}
AI Scientist~\cite{ai_scientist} and AI Scientist-v2~\cite{ai_scientist_v2} represent the closest competing approach to end-to-end scientific automation. The critical difference is the source of scientific contribution: AI Scientist's ideas originate from the LLM's parametric memory---essentially recombining existing knowledge---while ResearchEVO's discoveries emerge from principled evolutionary search over an algorithm space. This distinction matters because parametric-memory-based ideation is fundamentally limited to the training distribution, whereas evolutionary search can discover solutions that no human has previously proposed. AlphaEvolve~\cite{alphaevolve} shares the evolutionary discovery capability but produces no scientific documentation. Google's AI Co-Scientist~\cite{google_coscientist} generates hypotheses through multi-agent debate but neither implements them as optimized code nor produces papers. CycleResearcher~\cite{cycleresearcher} achieves impressive paper quality through RL-based writing optimization but lacks a principled discovery mechanism. ResearchEVO occupies the unique intersection of these capabilities.

\paragraph{On Depth vs.\ Breadth.}
ResearchEVO validates on two case studies rather than a large benchmark suite. We deliberately prioritize depth over breadth: each case study involves real scientific domains (not ML toy problems), rigorous statistical methodology (10 random seeds, bootstrap CI, paired sign tests, full ablation), and validation on real hardware data (QEC). The two domains---quantum computing and scientific computing---are sufficiently distinct to demonstrate cross-disciplinary generalization while allowing thorough investigation of each.

\section{Broader Impacts and Limitations}
\label{sec:impacts}

\paragraph{Broader Impacts.}
ResearchEVO takes a step toward democratizing scientific research by enabling researchers without deep expertise in adjacent domains to generate well-grounded initial research directions. A quantum physicist could use it to discover improved decoding strategies; the Writing Phase would automatically situate the discovery within the ML optimization literature. The end-to-end formulation also provides structural safeguards against fabrication: because all paper claims are anchored in actually evolved and evaluated code, the system cannot generate results that were not computationally verified. The sentence-level RAG with explicit citation verification further reduces hallucination risk.

\paragraph{Limitations.}
(1) The Evolution Phase inherits the computational cost scaling $O(T \cdot M \cdot C_{\text{LLM}})$ common to LLM-guided evolutionary approaches, which becomes prohibitive for very fine-grained optimization. (2) Despite RAG grounding, the LLM may occasionally mis-cite or over-generalize; generated papers require human review before submission. (3) The Writing Phase's quality depends partly on the \LaTeX\ template and rubric design, which require domain-specific tuning. (4) A full pipeline consumes thousands of LLM calls and several hours of wall-clock time. (5) The current design is a sequential pipeline; tighter coupling where writing feedback informs evolution is left to future work. (6) Generated papers have not yet undergone formal peer review. (7) We validate on two cross-disciplinary scientific problems; scaling to additional domains is a natural next step.

\section{Conclusion}
\label{sec:conclusion}

Many significant scientific discoveries have followed a two-stage pattern: first, blind exploration that yields an unexpected finding; then, retrospective analysis that explains why it works. ResearchEVO computationally instantiates this pattern through two dedicated phases---the Evolution Phase discovers algorithms purely by fitness through bi-dimensional co-evolution over functional and structural dimensions, and the Writing Phase retroactively explains the discoveries through literature-grounded analysis with sentence-level retrieval and explicit citation verification. To our knowledge, this is the first system to cover the full pipeline from problem specification to publication-ready paper end to end.

The power of this separation was demonstrated in both case studies. In QEC, the Evolution Phase blindly discovered topologically-aware edge reweighting factors that align with known---but previously unexploited---hardware physics; the Writing Phase then connected these factors to surface-code topology and noise characterization literature. In PINN, the evolution independently found a trust-region constraint and residual connections addressing orthogonal failure modes; the Writing Phase linked these to classical optimization theory and spectral bias research. In both cases, the discoveries were human-interpretable, and the generated papers contained zero fabricated citations.

ResearchEVO opens a productive direction at the intersection of evolutionary computation, LLM reasoning, and scientific methodology. Future work includes tighter phase coupling where writing analysis feeds back to guide evolution, multi-objective optimization over accuracy-efficiency Pareto fronts, extension to experimental sciences involving wet-lab protocols, and formal peer review validation of generated manuscripts.

\bibliographystyle{plain}
\bibliography{references}

\newpage
\appendix
\section{The EvoAny Open-Source Platform and Technical Lineage}
\label{app:lineage}

\begin{figure*}[h]
  \centering
  \includegraphics[width=\linewidth]{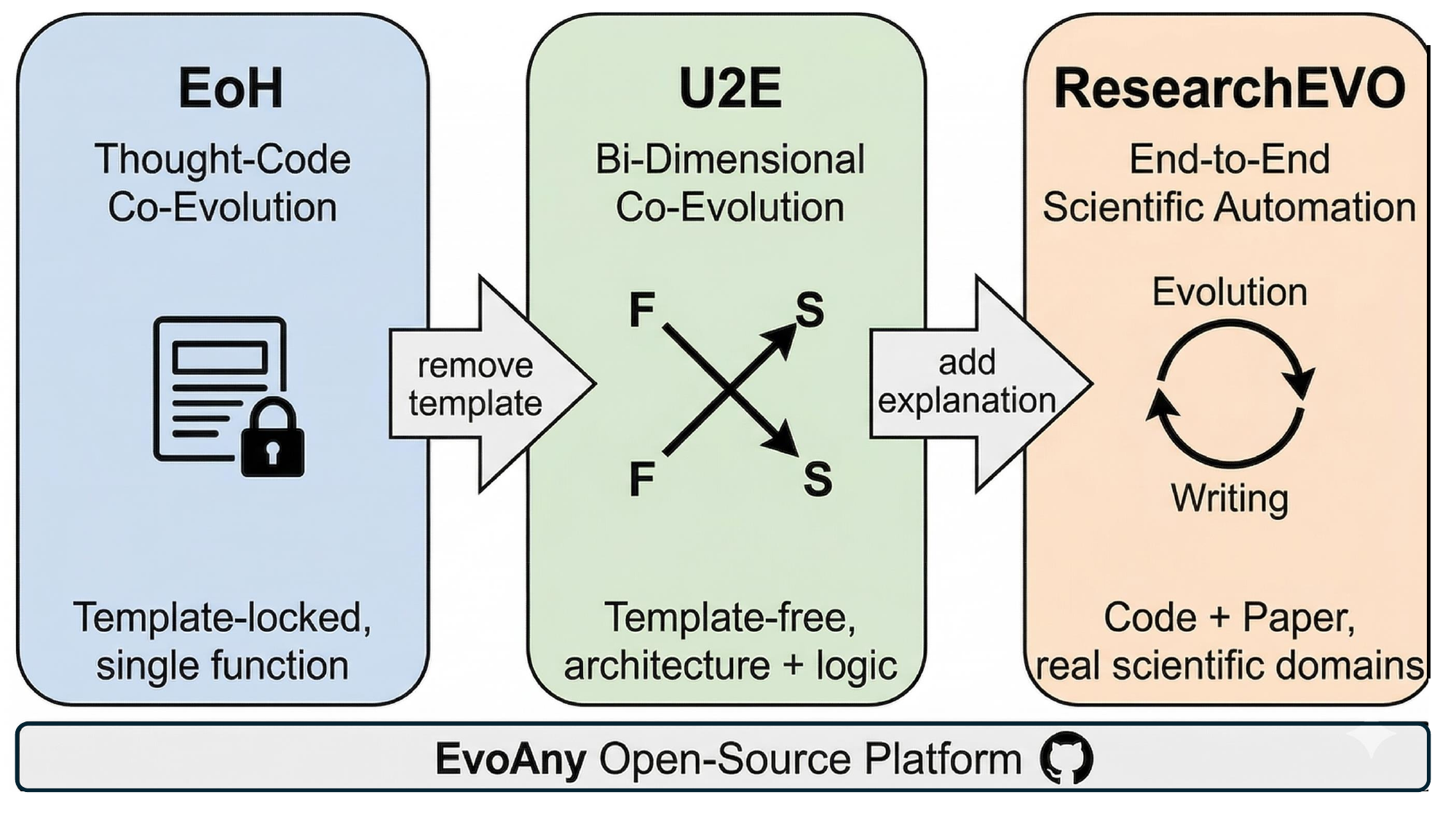}

  \caption{Technical evolution of the EvoAny platform. Each stage resolved the core bottleneck left by its predecessor, culminating in ResearchEVO's end-to-end scientific automation.}
  \label{tab:lineage}
\end{figure*}

ResearchEVO's Evolution Phase is implemented within \textbf{EvoAny}\footnote{\url{https://github.com/DataLab-atom/EvoAny}}, an open-source platform that unifies the technical lineage of LLM-driven algorithm evolution. EvoAny provides a single configurable codebase that supports multiple evolution strategies---from single-function optimization to full bi-dimensional co-evolution---and standardizes the problem definition interface so that new scientific domains can be onboarded with minimal engineering effort. Figure~\ref{tab:lineage} summarizes how this platform consolidates three stages of technical development.

\subsection{EoH: Thought-Code Co-Evolution}

EoH~\cite{liu2024evolution} introduced the foundational idea of co-evolving a natural-language ``thought'' (describing the algorithmic strategy) alongside the executable code implementation. By letting the LLM simultaneously serve as crossover and mutation operator, EoH achieved three orders of magnitude improvement in sample efficiency over prior approaches such as FunSearch~\cite{funsearch}. However, EoH requires a human-provided algorithm template: only a single pre-identified function within this fixed template is subject to evolution. The algorithm's overall architecture---control flow, module organization, data structures---remains locked, limiting achievable innovation to incremental refinements within a fixed framework.

\subsection{U2E: Template-Free Bi-Dimensional Co-Evolution}

U2E~\cite{u2e} directly addressed the core limitations of EoH through three extensions. First, it eliminated template dependence entirely: the LLM generates complete algorithm code from natural language problem descriptions, expanding the search space from ``functions within a template'' to ``the full algorithm design space.'' Second, it introduced automatic bottleneck identification, removing the need for human specification of which functions to evolve. Third, and most critically, it introduced bi-dimensional co-evolution---simultaneously optimizing both the functional dimension ($\mathcal{F}$, the internal logic of individual modules) and the structural dimension ($\mathcal{S}$, the overall algorithm architecture). This second dimension allows the evolutionary process to discover entirely new algorithmic paradigms rather than merely refining existing ones. U2E demonstrated this capability on combinatorial optimization problems, where the evolution trajectory from generation 0 to generation 7 showed not just improved function implementations but fundamentally restructured algorithm architectures.

\subsection{ResearchEVO: Completing the Scientific Pipeline}

ResearchEVO extends U2E in two directions. First, it introduces domain-adaptive sandbox evaluation that returns structured error feedback (stack traces, partial outputs, diagnostic messages) beyond scalar fitness scores, enabling the bi-dimensional co-evolution engine to operate on scientific problems (quantum error correction, physics-informed neural networks) where execution failures carry rich diagnostic information. Second, and more fundamentally, ResearchEVO completes the discover-explain pipeline by adding the Writing Phase: a three-stage pipeline that takes the discovered algorithm $W^*$ and autonomously produces a complete research paper with sentence-level RAG and explicit anti-hallucination verification. This transforms the output from ``a piece of optimized code'' into ``a piece of optimized code together with a complete scientific paper explaining it''---covering the full scientific value chain from discovery through explanation to documentation.

All three stages are available within the EvoAny platform. Researchers can select the appropriate evolution strategy (EoH-style single-function evolution, U2E-style bi-dimensional co-evolution, or the full ResearchEVO pipeline with writing) depending on their needs, and extend the platform to new scientific domains by providing a problem description, evaluation oracle, and seed bibliography.

\section{Comparison with AlphaEvolve}
\label{app:alphaevolve}

AlphaEvolve~\cite{alphaevolve} is a concurrent system that also performs LLM-guided code evolution at scale. While both systems move beyond single-function template-based evolution, they occupy fundamentally different positions in the design space. We note that the bi-dimensional co-evolution workflow used in ResearchEVO was established in U2E~\cite{u2e}, which predates AlphaEvolve.

\subsection{Structured 2D Search vs.\ Flat Diff Search}

AlphaEvolve generates code patches (diffs) that may simultaneously modify function logic and control flow, but the system does not distinguish between these two types of modification. It cannot explicitly decide ``this iteration should restructure the architecture'' versus ``this iteration should refine a specific function.'' ResearchEVO decomposes the search into an explicit functional dimension and structural dimension, enabling targeted switching: when functional optimization reaches diminishing returns, the system recognizes that the architecture has become the binding constraint and shifts to structural evolution. This structured decomposition, combined with reflective feedback within each dimension, achieves effective search with far smaller computational budgets than AlphaEvolve's MAP-Elites approach, which relies on large-scale random diversity to occasionally stumble upon architectural changes.

\subsection{Human-Interpretable Outputs vs.\ Black-Box Outputs}

AlphaEvolve's most celebrated result---the first improvement over Strassen's matrix multiplication algorithm in fifty-six years---produced a set of numerical matrix coefficients whose optimality no human can explain. Similarly, its TPU layout and data-center scheduling optimizations are black-box improvements. By contrast, ResearchEVO's discoveries are human-interpretable: DOA-MWPM consists of four binary topological properties with intuitive scaling factors that physicists can understand, verify, and build upon; ResLRA-PINN consists of a trust-region constraint and residual connections that optimization theorists can immediately relate to classical methods. This interpretability is not accidental but a consequence of the structured 2D search: because functional and structural changes are tracked separately, the evolution trajectory itself is attributable---each improvement can be identified as either a logic refinement or an architectural restructuring.

\subsection{End-to-End Scientific Documentation vs.\ Code-Only Output}

AlphaEvolve produces optimized code as its sole output. The scientific interpretation, contextualization, and documentation of its discoveries must be performed manually by Google's research team. ResearchEVO's Writing Phase automatically generates a complete research paper that explains the discovered algorithm, connects it to existing literature through sentence-level RAG, designs and executes validation experiments, and verifies every citation against an indexed database. This covers the full scientific value chain from discovery through explanation to dissemination.

\subsection{Accessibility vs.\ Industrial-Scale Requirements}

AlphaEvolve uses a proprietary ensemble of Gemini Flash and Pro models running on Google's internal infrastructure. It is not open-source, and reproducing its results requires computational resources available only to a handful of industrial labs. ResearchEVO operates with publicly available LLMs (GPT-4o), a population size of 10, and completes a full run---including both evolution and paper generation---in several hours on standard hardware. The entire evolution engine is open-sourced through the EvoAny platform, making the approach accessible to academic researchers and reproducible without industrial-scale resources.

\subsection{Where AlphaEvolve Excels}

AlphaEvolve holds clear advantages in raw scale and mathematical impact. Its validation on Google production systems (TPU layout optimization, data-center scheduling) and its breakthrough on Strassen's algorithm represent achievements that ResearchEVO's two academic case studies do not match in magnitude. AlphaEvolve's MAP-Elites diversity maintenance provides more structured population diversity than ResearchEVO's simpler rank-based selection. The two systems ultimately address different questions: AlphaEvolve asks ``how to find better code at industrial scale,'' while ResearchEVO asks ``how to automate the full scientific research cycle from discovery to documentation.''

\end{document}